\documentclass[runningheads]{llncs}

\usepackage{color}
\usepackage{graphicx}
\usepackage{amsmath}
\usepackage{wrapfig}
\usepackage{booktabs} 
\usepackage{mathtools}
\usepackage{url}
\usepackage[utf8x]{inputenc}

\DeclareMathOperator*{\argmax}{arg\,max}

\newcommand{\eqdef}{=\vcentcolon}

\newcommand\il[1]{\langle #1 \rangle}

\def\our{\textsc{NetCut}}

\begin{document}
\title{Finding the Optimal Network Depth in Classification Tasks}
%
%
\author{Bartosz W\'o{}jcik \and Maciej Wo\l{}czyk \and Klaudia Ba\l{}azy \and Jacek Tabor}
\authorrunning{B. W\'o{}jcik et al.}
%
\institute{GMUM, Jagiellonian University, Krak\'o{}w, Poland}
\maketitle              
\begin{abstract}

We develop a fast end-to-end method for training lightweight neural networks using multiple classifier heads. By allowing the model to determine the importance of each head and rewarding the choice of a single shallow classifier, we are able to detect and remove unneeded components of the network. This operation, which can be seen as finding the optimal depth of the model, significantly reduces the number of parameters and accelerates inference across different hardware processing units, which is not the case for many standard pruning methods. We show the performance of our method on multiple network architectures and datasets, analyze its optimization properties, and conduct ablation studies. 

\end{abstract}

\section{Introduction}

Although deep learning methods excel at various tasks in computer vision, natural language processing, and reinforcement learning, running neural networks used for those purposes is often computationally expensive \cite{zhang2016understanding,yang2019xlnet}. At the same time, many real-life applications require the models to run in real-time on hardware not specialized for deep learning techniques, which is infeasible for networks of this scale. Therefore, it is crucial to find methods for reducing both memory requirements and inference time without losing performance. In response to this problem, the field of model compression emerged, with pruning being one of the most important research directions \cite{cheng2017survey,choudhary2020comprehensive}.

Pruning methods that rely on removing parameters from a large trained network based on some metric (e.g. magnitude of the weights or approximated loss on removal) are very effective at reducing the size of the model. Nonetheless, it is unclear how good those methods are at decreasing the inference time of the models.

\begin{figure}[h!]
\centering 
\includegraphics[width=\linewidth]{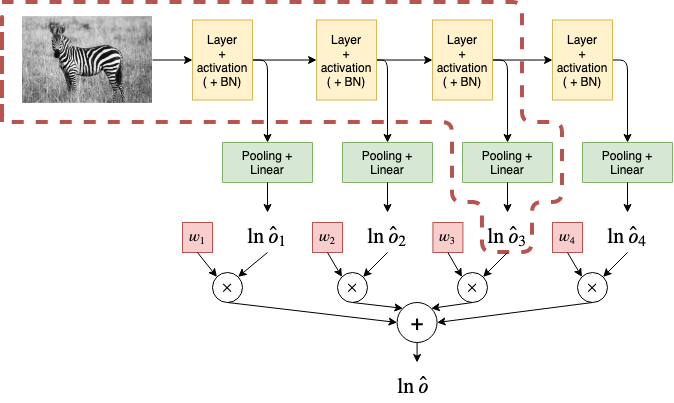}
\caption{We attach classifier heads (green blocks) to the base feed-forward network (yellow blocks) and then use their combined outputs as the final prediction of the model. The output of each classifier is weighted by its weight $w_k$, with $\sum_k w_k = 1$. During training, we force the model to converge to a single $w_l \approx 1$, and thus we can remove all layers after $l$ and all heads other than the $l$-th one to obtain a compressed network (red dashed line).}
\label{fig:architecture}
\end{figure} 

Unstructured pruning methods excel at creating sparse representations of overparameterized networks without losing much performance by removing single connections \cite{gale2019state}. However, special hardware is required in order to take advantage of the sparsity and decrease the inference time \cite{turner2018characterising}. Structured pruning aims to solve this problem by removing whole channels instead of single connections, which accelerates the network even when running on general-purpose hardware. At the same time, recent research suggests that the acceleration-performance trade-off in networks obtained via structured pruning is worse than in shallower networks trained from scratch \cite{crowley2018closer}. 

The approach of minimizing the depth of the network can be also motivated from a biological perspective. Recent research of functional networks in the human brain shows that learning a task leads to a reduction of communication distances \cite{heitger2012motor,sami2013graph}. Thus, it may be beneficial to try to simulate the same phenomenon in the case of artificial neural networks by decreasing the length of the paths in artificial neural networks (i.e. the depth) instead of minimizing the width (i.e. the number of connections in each layer).

Inspired by those findings, we introduce \our{}, a quick end-to-end method for reducing the depth of the network in classification tasks. We add a classifier head on top of each hidden layer and use their combined predictions as the final answer of the model, where the influence of each head is a parameter updated using gradient descent. The basic scheme of the model is presented in Fig. \ref{fig:architecture}.

Our main theoretical contribution is a method of aggregating outputs from individual classifier heads, which enforces choosing only one head during the training and simplifies the model in the process. We show that combining the logarithm of probabilities instead of probabilities themselves encourages the network to choose a model with a single classifier head. This method also allows us to avoid the numerical instability linked to using a standard softmax layer. Finally, we introduce a regularization component to the loss function that simulates the time it takes the network to process the input. After the network converges to a single classifier head, we cut out the resulting network without a noticeable performance drop.

We test \our{} in multiple settings, showing stable performance across different models, including experiments on highly non-linear network architectures obtained by generating random graphs. Our findings show that the resulting cut-out shallow networks are much faster than the original one, both on CPU and GPU while maintaining similar performance. Finally, we perform an extensive analysis of the optimization properties of our method. We examine the directions of gradients of each classifier, different initialization schemes for the weight layers and the effects of poor starting conditions.


\section{Related Work}
%

Our method can be seen as the pruning of whole layers instead of single connections or neurons. Similar ideas have been investigated in the past. Wen et al. \cite{wen2016learning} introduced depth-regularization in ResNets by applying group Lasso on the whole layers. Huang \& Wang \cite{huang2018data} explored a similar idea from the perspective of forcing the output of the entire layer to fall to $0$ instead of individual weights in the layer. However similar, those approaches are only applicable for architectures with residual connections, where \our{} works for any chosen feed-forward network.



Attaching multiple output heads to a base neural network, which is a central idea in our approach, has been previously exhibited in many works. Considering the publications most strongly connected to our work, Huang et al. \cite{huang2017multi} used a multi-scale dense network in order to allow for the dynamical response time of the model, depending on the difficulty of the input sample. Zhang et al. \cite{zhang2019scan} extended this idea by introducing knowledge distillation between classifiers in the model to improve the performance of earlier heads and using custom attention modules in order to avoid negative interference in the base of the network. However, since those methods focus on dynamical resource management, it is not clear how to use them to obtain smaller, quicker networks, which is the goal of our work.

We would like to point out that our approach is in many ways orthogonal to the methods presented in this section, and as such may be combined with other techniques to further improve the results. For example, enforcing knowledge distillation between the classifiers in our method could improve the accuracy of the earlier layers as it is shown in \cite{zhang2019scan}. Additionally, further parameter reduction could be obtained by applying pruning methods on the cut-out network produced by \our{}.

\section{Theoretical Model}
In this section we describe our model. Subsection~\ref{subsection:multi_head} presents a basic approach for using a multi-head network for classification. Subsection~\ref{subsection:logaggregation} introduces our novel way of combining probabilities from classifiers. In Subsection~\ref{subsection:time_regularization} we describe our regularization scheme and show how compression is achieved.

\subsection{Basic multi-head classification model}
\label{subsection:multi_head}
\our{} considers an overparameterized feed-forward neural network architecture, which we want to train for a classification task with the final goal of reducing the computational effort during inference.

We modify the original network by adding classifier heads on top of hidden layers, as it is presented in Fig. \ref{fig:architecture}. Thus, for every input, we get multiple vectors $\hat{o}_1, \ldots, \hat{o}_n$, where $n$ is the number of classifiers, and by $\hat{o}_k^{(i)}$ we denote the probability of the $i$-th class given by the $k$-th classifier.
This modification is non-invasive as the architecture of the core of the network remains unchanged, and the classifier heads have very few parameters compared to the original network.

The classifiers are then trained together by combining the probabilities $\hat{o}_k$ in order to produce the final prediction of the network $\hat{o}$, which is then used as an input to the cross-entropy loss
\begin{equation*}
    L_{\mathrm{class}}(y, \hat{o}) = -\sum_i y^{(i)} \log \hat{o}^{(i)},
\end{equation*}
where $y$ is the true label represented as a one-hot vector.

In order to allow the model to choose which classifiers are useful for the task at hand, we introduce classifier head importance weights $w_k$, with $w_k \geq 0$ and $\sum_k w_k = 1$. The weights are parameters of the network, and the model updates them during the gradient descent step.

Then, the basic way of aggregating the outputs of each classifier would be to use a weighted average of the probabilities
\begin{equation*}
    \label{softmax_averaging}
    \hat{o}^{(i)} = \sum_k w_k \hat{o}^{(i)}_k.
\end{equation*}
Therefore, the resulting model may be seen as an ensemble of classifiers of different depth. 

\subsection{\our{} aggregation scheme}
\label{subsection:logaggregation}
The presented way of aggregating the outputs of the classifiers does not encourage the network in any way to choose just one classifier. This reduction is crucial for the proper compression of the model. In order to enforce it, we propose a novel way of combining the outputs of multiple classifiers, which heavily rewards choosing just one layer:
$$ 
\hat{o}^{(i)} = \exp \sum_k w_k \ln\hat{o}^{(i)}_k.
$$ 

The resulting vector $\hat{o}$ does not necessarily represent a probability distribution as its elements $\hat{o}^{(i)}$ do not always sum up to $1$:
\begin{equation*}
\label{eq:jensen}
\begin{split}
\sum_i \hat{o}^{(i)} = \sum_i \exp\sum_k w_k \ln \hat{o}_k^{(i)} &\leq \sum_i \exp\ln \sum_k w_k \hat{o}_k^{(i)} = 1, 
\end{split}
\end{equation*}
where the second transition comes from Jensen's inequality.

One can see that the cross-entropy loss is minimized when $\hat{o}^{(i)} = 1$ for the correct class $i$, which is possible only when equality in the above formula holds. This is the case when $w_l = 1$ for some $l$ and $w_k = 0$ for all $k \neq l$, since
\begin{equation*}
    \sum_i \hat{o}^{(i)} = \sum_i \exp \ln \hat{o}_l^{(i)} = 1.
\end{equation*}

Other equality conditions for Jensen's inequality cannot be satisfied in our setting, since the logarithm is not an affine function and it is impossible for every $\hat{o}_k$ to be equal because of the inherent noise of the training procedure. Thus, in order to reduce the value of the cross-entropy loss function in our approach, the method will strive towards leaving only one nonzero weight $w_l = 1$. This in turn is equivalent to choosing a single-head subnetwork which can be easily cut out from the starting network without losing performance.

Concluding, we arrive at the following theorem: 
\begin{theorem}
Suppose that the $l$-th classifier head of \our{} network obtains a perfect classification rate, i.e. it always assigns probability $1$ to the correct class. By minimizing the loss function of the model with our aggregation scheme, we will arrive at $w_l = 1$, and $w_k = 0$ for all $k \neq l$.
\end{theorem}
Observe that the same result is not valid for the basic aggregation scheme.

Indeed in practice, as we show in Section~\ref{sec:experiments}, combining the log probabilities of each layer causes the network to quickly focus on only one of the classifiers. The same is not the case for weighting the probabilities directly, where the network spreads the mass across multiple head importance weights, as we show in Section~\ref{subsection:weightnormalization}. Such a subnetwork cannot be then easily extracted.

We emphasize the positive numerical properties of this approach. In order to combine the predictions by directly weighting the probabilities outputted by each layer, we would have to calculate the softmax function for each classifier, which could potentially introduce numerical instabilities. Meanwhile, in our approach we can avoid computing the softmax function by directly computing the log probabilities. This technique, known in the computer science community as the log-sum-exp trick is less prone to numerical under- and overflows \cite{blanchard2019accurate}. 

\subsection{Time-regularization and model compression}
\label{subsection:time_regularization}
Our goal is to train a network which will be significantly faster during inference, i.e. we want to minimize the time of processing a sample. However, the time the network takes to process a sample is not a differentiable object and thus cannot be used directly to update the network with standard gradient descent methods.

In order to approximate the inference time, we introduce a surrogate differentiable penalty based on the number of layers in the network:
$$
L_{\mathrm{reg}} = \sum_k w_k k.
$$

For scaling the regularization cost, we introduce the hyperparameter $\beta$, giving the final loss function of the model

$$
L = L_{\mathrm{class}} + \beta L_{\mathrm{reg}}.
$$

After training, we can reduce the size of the network by using only the layer chosen by the model, as given by $l = \argmax_k w_k$. The trained network can be then easily compressed by cutting out all the layers with indices larger than $l$ and removing all the classifier heads other than $l$-th one, as presented in Fig. \ref{fig:architecture}.

In theory, the "cut-out" network may perform worse than the original one, because of the lack of influence of the removed classifiers. However, in practice, we show that the final weight $w_l \approx 1$ at the end of the training, and the influence of the rest of the layers is negligible. This is not necessarily the case for weighting schemes other than our log softmax approach.

\section{Experiments}
\label{sec:experiments}

In this section we show the performance of \our{} on multiple models, highlighting its applicability in various settings. The following subsections describe experiments on standard vision architectures (traditional CNNs, ResNets, and fully connected networks). Then, Subsection~\ref{sec:graphs} presents experiments in which we apply \our{} to randomly generated graph-based networks.

\begin{figure}[h!]
\centering 
\includegraphics[width=0.7\textwidth]{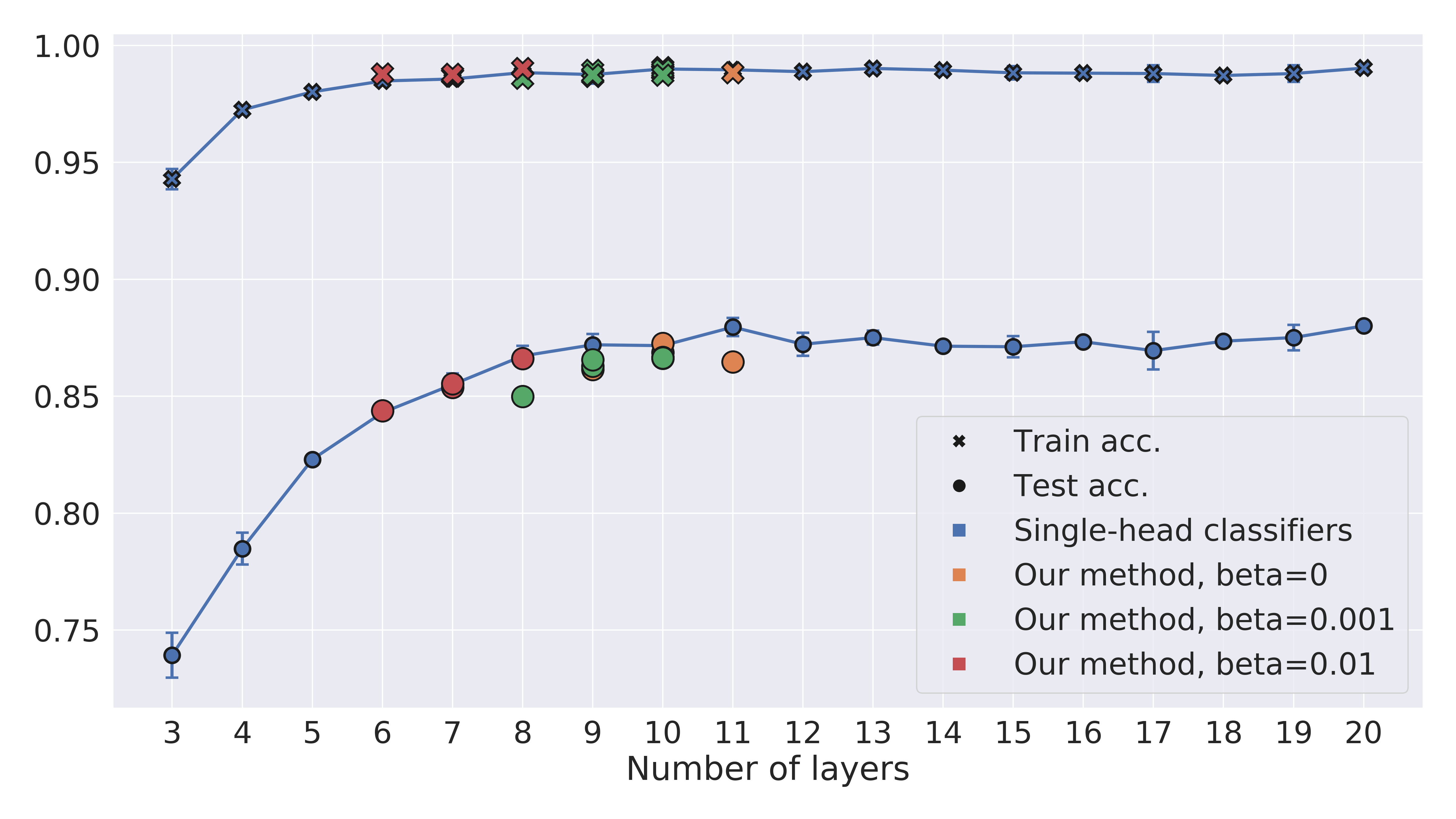}
\caption{Accuracy of \our{} on a standard convolutional architecture trained on CIFAR-10. Results for the single-head baseline networks, representing all the subnetwork our network can pick, are averaged over 3 runs and plotted with error bars. As each run of our method can converge to a different layer, we plot them separately. Observe that \our{} shrinks the starting $20$-layer network without a significant accuracy drop.}
\label{fig:beta_baselines_cifar_conv_bn}
\end{figure}

The head importance weights $w_k$ are initialized uniformly in each experiment, and the depth-regularization coefficient $\beta$ is set to $0$ unless specifically noted otherwise. 
We published the code used for conducting the experiments on GitHub: \url{https://github.com/gmum/classification-optimal-network-depth}.

\subsection{Standard CNNs}
\label{sec:standard_cnns}

\begin{figure}[h!]
\begin{minipage}{.32\textwidth}
  \centering
  \includegraphics[height=3.9cm,keepaspectratio]{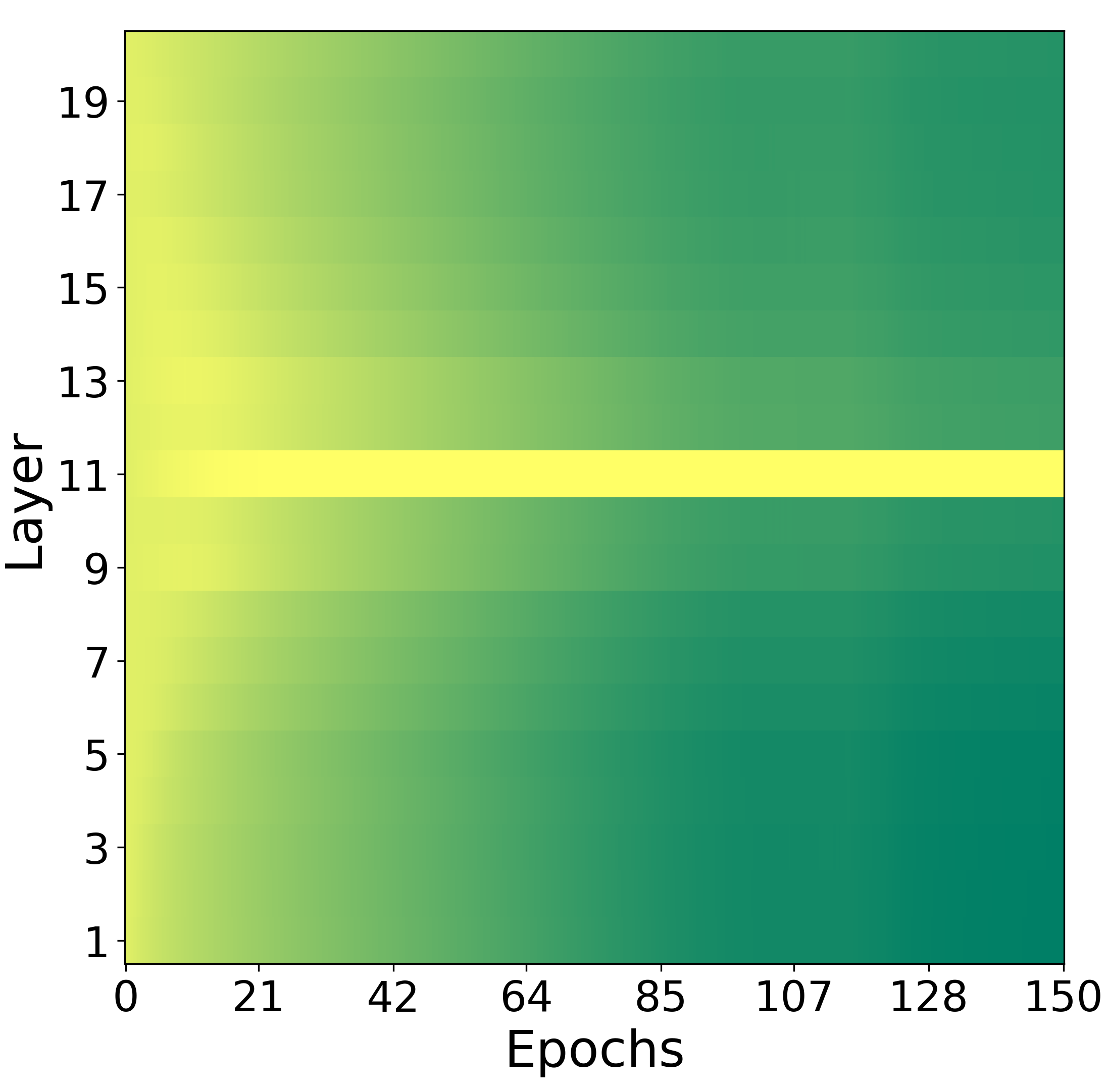}
  (a) $\beta = 0$
  \label{fig:beta_baselines_weights_a}
\end{minipage}%
\begin{minipage}{.32\textwidth}
  \centering
  \includegraphics[height=3.9cm,keepaspectratio]{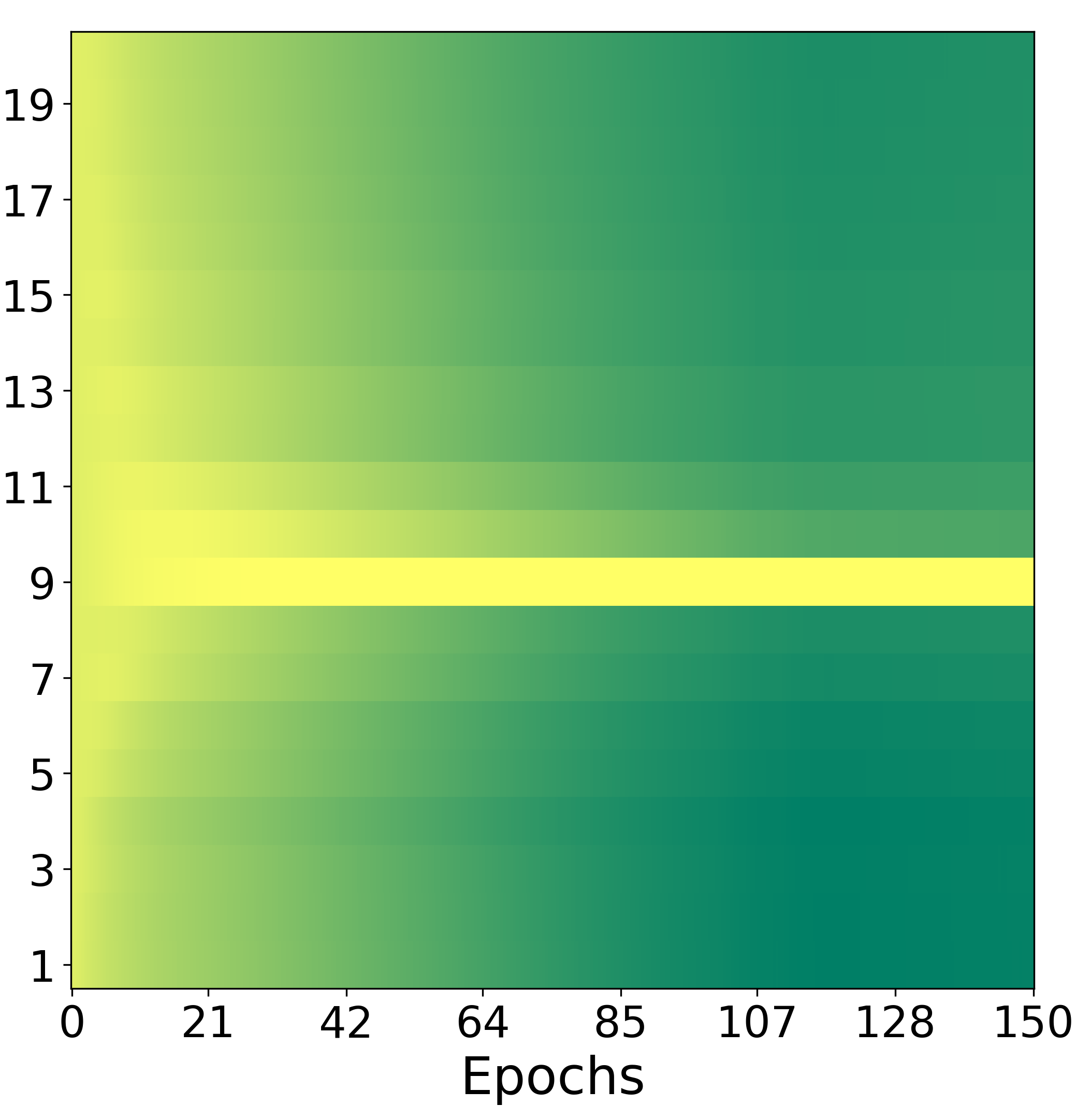}
  (b) $\beta = 0.001$
  \label{fig:beta_baselines_weights_b}
\end{minipage}%
\begin{minipage}{.32\textwidth}
  \centering
  \includegraphics[height=3.9cm,keepaspectratio]{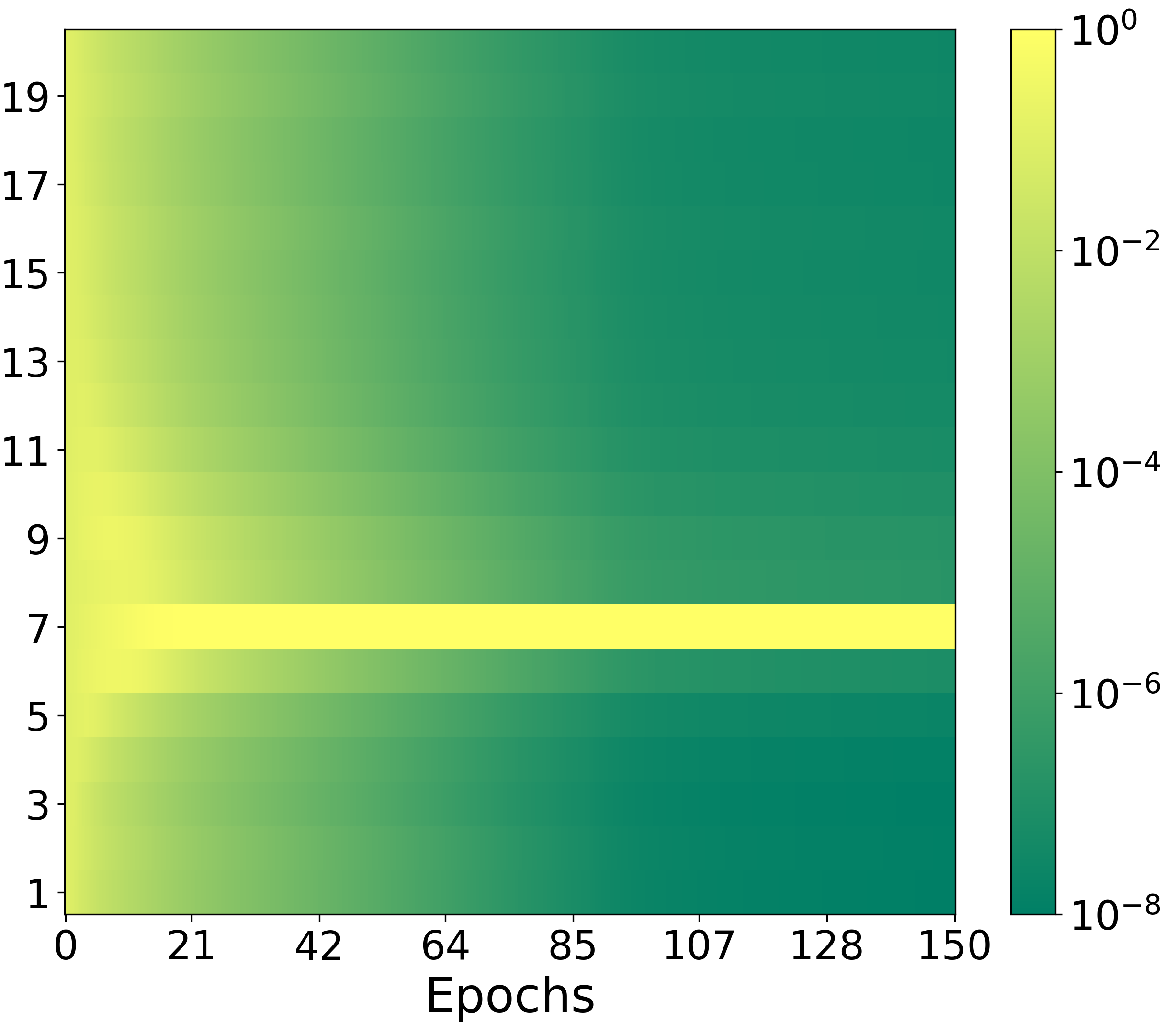}
  (c) $\beta = 0.01$
  \label{fig:beta_baseliens_weights_c}
\end{minipage}%
\caption{Classifier head importance weights $w_k$ training progress for different $\beta$ values for a convolutional architecture trained on CIFAR-10. Colors in each row represent a change of a single weight through time. We use a logarithmic color map (as shown on the scale on the right) in order to show the differences in smaller values. The results show that the method decides on a layer very quickly and that by increasing the $\beta$ hyperparameter we can encourage the model to select a smaller subnetwork. }
\label{fig:beta_baselines_weights}
\end{figure}

\begin{wrapfigure}{r}{0.50\textwidth}
\centering
\includegraphics[width=0.57\textwidth]{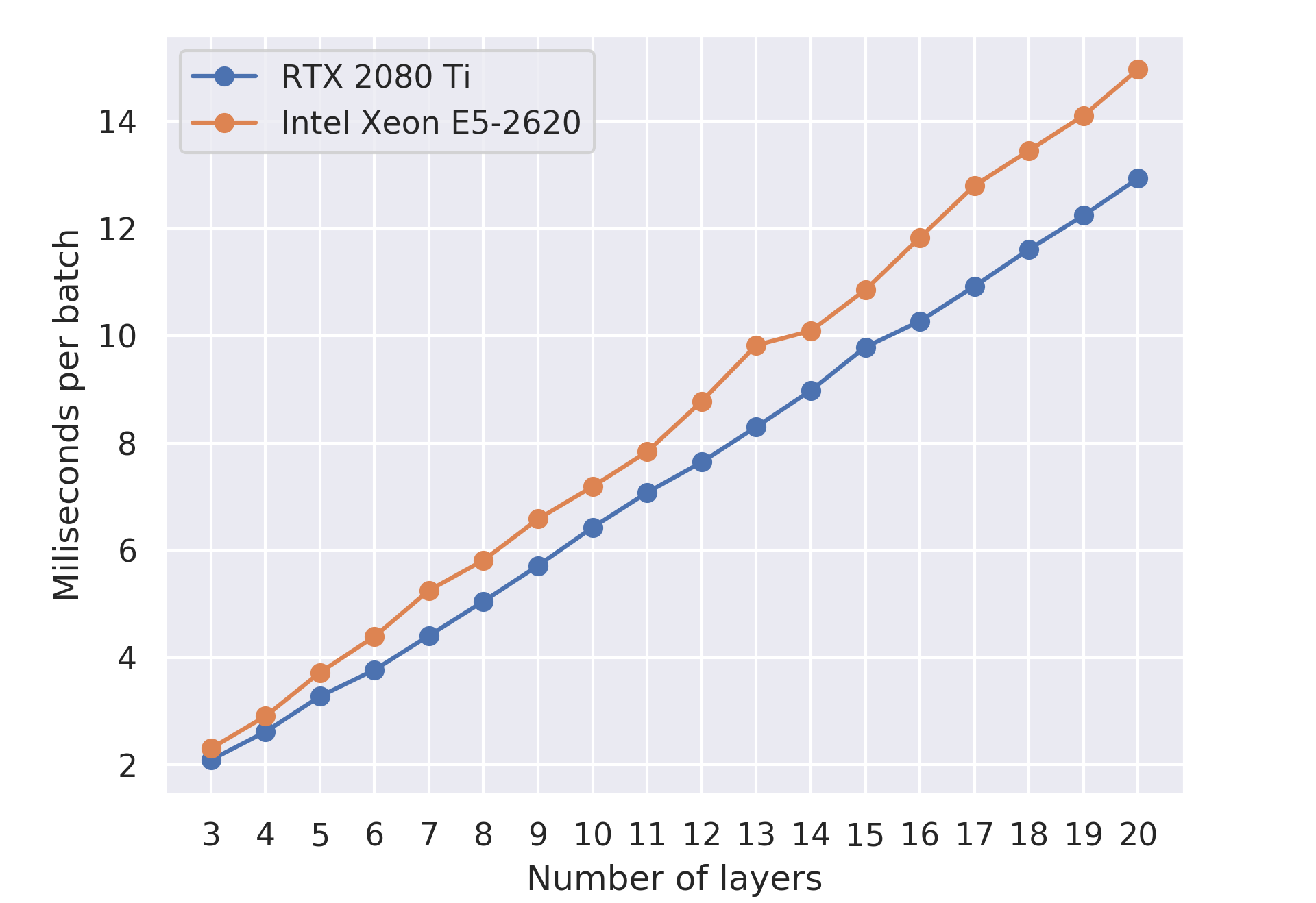}
\caption{Processing time of a single batch. The batch size is $1$ for CPU and $64$ for GPU. Decreasing the number of layers linearly reduces the inference time.}
\label{fig:inference_time}
\end{wrapfigure}

We begin testing our approach with a standard convolutional neural network with $20$ base convolutional layers trained on the CIFAR-10 dataset. Each convolutional layer consists of $50$ filters of size $5 \mathrm{x} 5$. We use batch normalization and the ReLU activation function. Each classifier head consists of a global max pooling layer and a fully-connected classification layer. We train the network for $150$ epochs using the Adam optimizer, batch size set to $128$, with basic data augmentation (random crop, horizontal flip), and without explicit regularization other than the optional time-regularization described in Subsection~\ref{subsection:time_regularization}.

To evaluate the performance of \our{}, we run the experiment $5$ times for each hyperparameter setting. As a baseline, we compare ourselves to all the possible subnetworks our method could pick during the training. To do so, we train $n$-layer single-head subnetworks of our starting network, with $n$ ranging from $3$ to $20$. In Fig. \ref{fig:beta_baselines_cifar_conv_bn} we report the train and test accuracy scores of those baseline single-head networks along with the results of our method with different $\beta$ hyperparameter values.

The results show that our approach is able to find a working subnetwork of much lower complexity, but similar accuracy. We can also control the shallowness of the network by changing the $\beta$ hyperparameter value, which determines the magnitude of the model complexity penalty in the loss function. In Fig. \ref{fig:beta_baselines_weights} we show how head importance weights $w_k$ change for a randomly chosen run from Fig. \ref{fig:beta_baselines_cifar_conv_bn} for each $\beta$ value. The regularization effect is evident from the very start of the learning process.

To show that the networks obtained by \our{} are significantly faster, we check the inference speed of single-head networks of varying depth and report the results in Fig. \ref{fig:inference_time} for CPU and GPU. As expected, the time needed to process a single batch increases linearly with the number of layers. This means that networks obtained by our method can increase the inference speed up to $2.85$ times compared to the starting network, with little to no performance drop.

\subsection{ResNets}

\begin{table}[]
\centering
\resizebox{0.75\textwidth}{!}{%
\begin{tabular}{c|ccccc|c}
\toprule
Reg. coef. $\beta$     & 0.000   & 0.002 & 0.004 & 0.008 & 0.010  & Original  \\
\midrule
Chosen block    & 41    & 28    & 24    & 10    & 7     & 54       \\
Accuracy & 89.84\% & 88.20\% & 87.50\%  & 80.33\% & 78.45\% & 91.6\%     \\
\bottomrule
\end{tabular}
}
\vspace{0.3cm}
\caption{ResNet-110 results for different $\beta$ values. The chosen head (block) and final accuracy are shown and compared to the original ResNet-110 architecture results. By changing $\beta$ we can directly balance the compression-performance trade-off.}
\label{tab:resnet-results}
\end{table}

\begin{figure}[h!]
\centering 
\begin{minipage}{.32\textwidth}
  \centering
  \includegraphics[height=3.9cm,keepaspectratio]{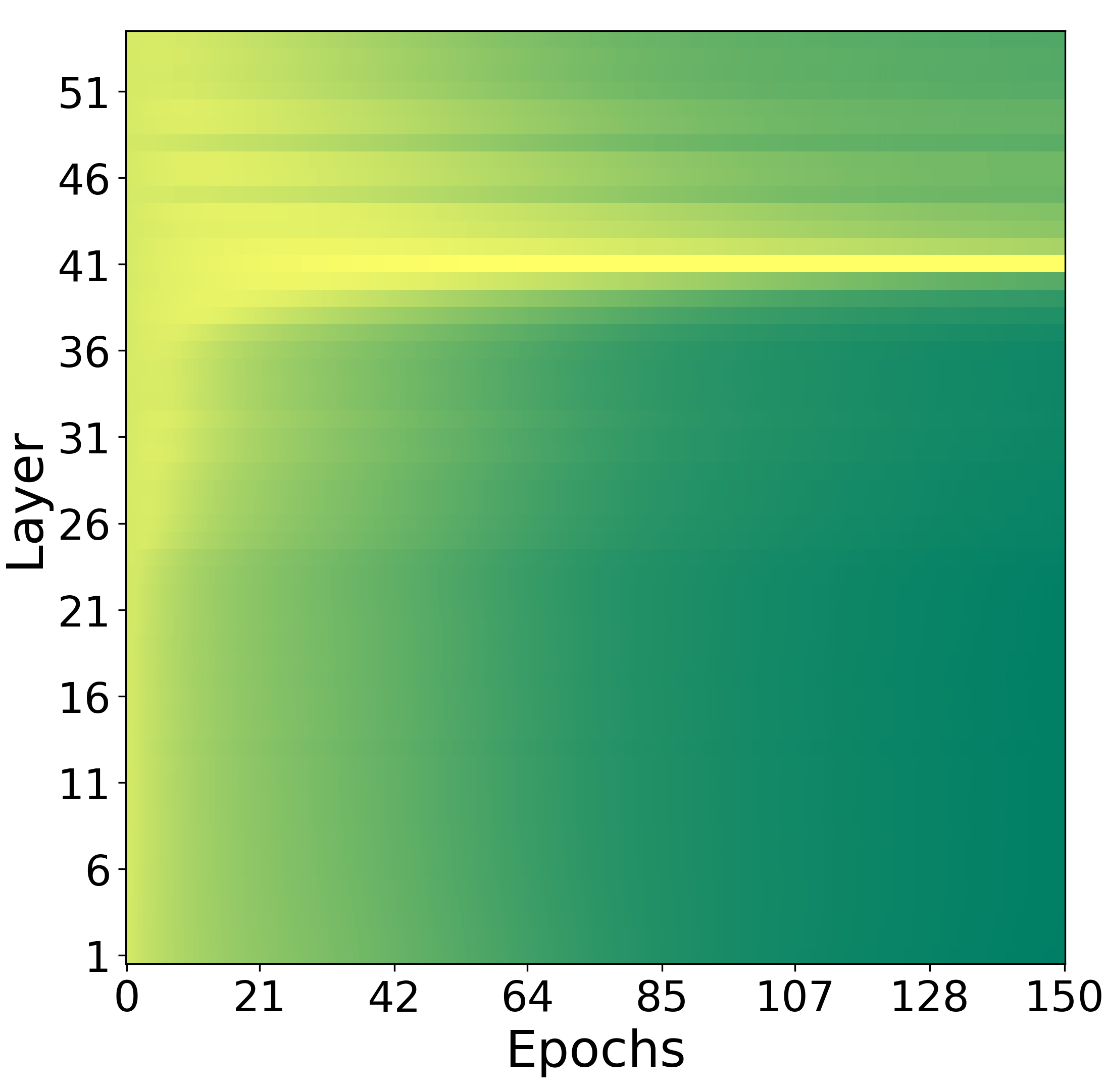}
  (a) $\beta = 0.0$
  \label{fig:beta_weights_resnet_0}
\end{minipage}
\begin{minipage}{.32\textwidth}
  \centering
  \includegraphics[height=3.9cm,keepaspectratio]{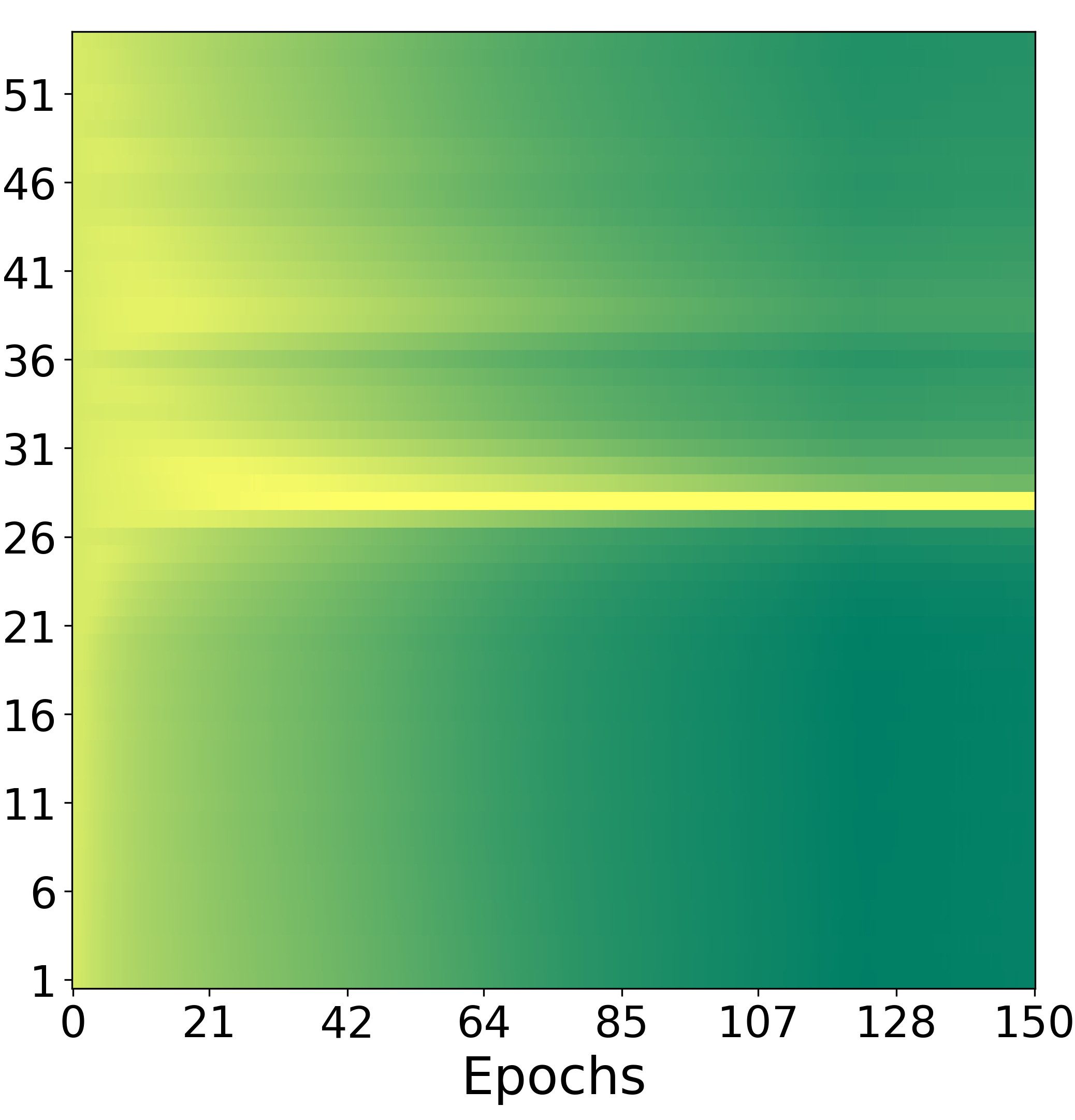}
  (b) $\beta = 0.002$
  \label{fig:beta_weights_resnet_2}
\end{minipage}
\begin{minipage}{.32\textwidth}
  \centering
  \includegraphics[height=3.9cm,keepaspectratio]{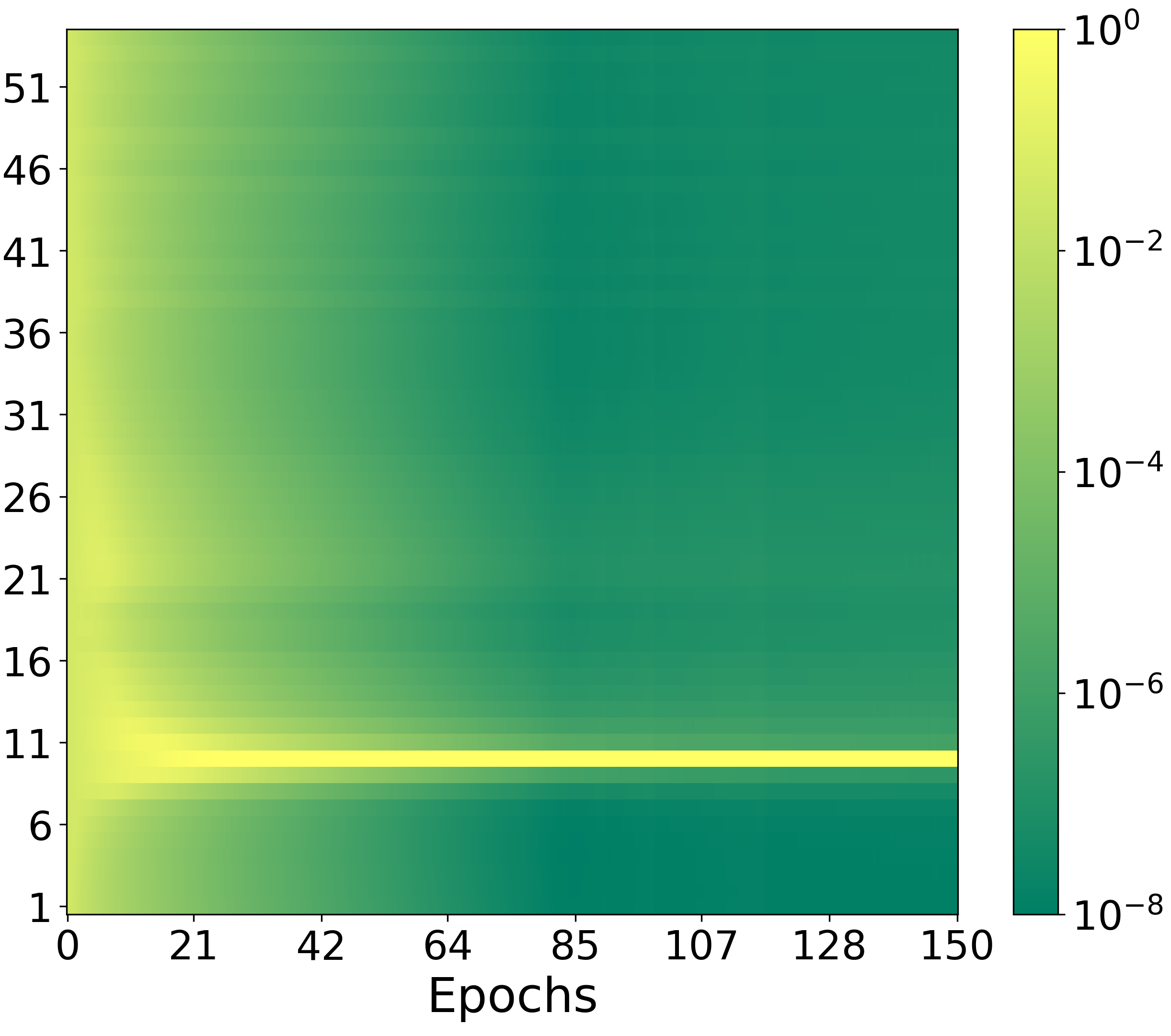}
  (c) $\beta = 0.008$
  \label{fig:beta_weights_resnet_8}
\end{minipage}
\caption{Head importance weights training progress for ResNet-110. Even with no regularization ($\beta = 0$) the network decreases the depth by almost 20\% and with higher regularization we can obtain a network that is ten times smaller.}
\label{fig:beta_weights_resnet}
\end{figure}

Subsequently, we test \our{} on ResNet-110 \cite{he2016deep} for CIFAR-10 by attaching a classification head after every block. Each head consists of a global average pooling layer and a fully-connected layer. We do not change any other architecture hyperparameters.
We present the results in Tab. \ref{tab:resnet-results}, which show the compression-performance trade-off for different values of the $\beta$ hyperparameter.

We have observed that for $\beta=0$ our model always chooses one of the final classifier heads, even if we modify the network and use a larger number of blocks. This is not the case for the standard CNN architectures studied in the previous subsection. We hypothesize that this effect is caused by the presence of residual connections, which encourage iterative refinement of features and thus make adding more layers preferable \cite{jastrzkebski2017residual}. 


\subsection{Fully Connected networks}

To cover a wider range of model types, we also test \our{} on simple fully-connected (FC) networks trained on MNIST and CIFAR-10 datasets. The width of every layer is $200$ for MNIST and $1000$ for CIFAR-10. Every head has the same architecture, consisting of a single fully-connected classification layer. We do not use batch normalization for this experiment.

\begin{figure}[h!]
\centering 
\begin{minipage}{.5\textwidth}
  \centering
  \includegraphics[width=\linewidth]{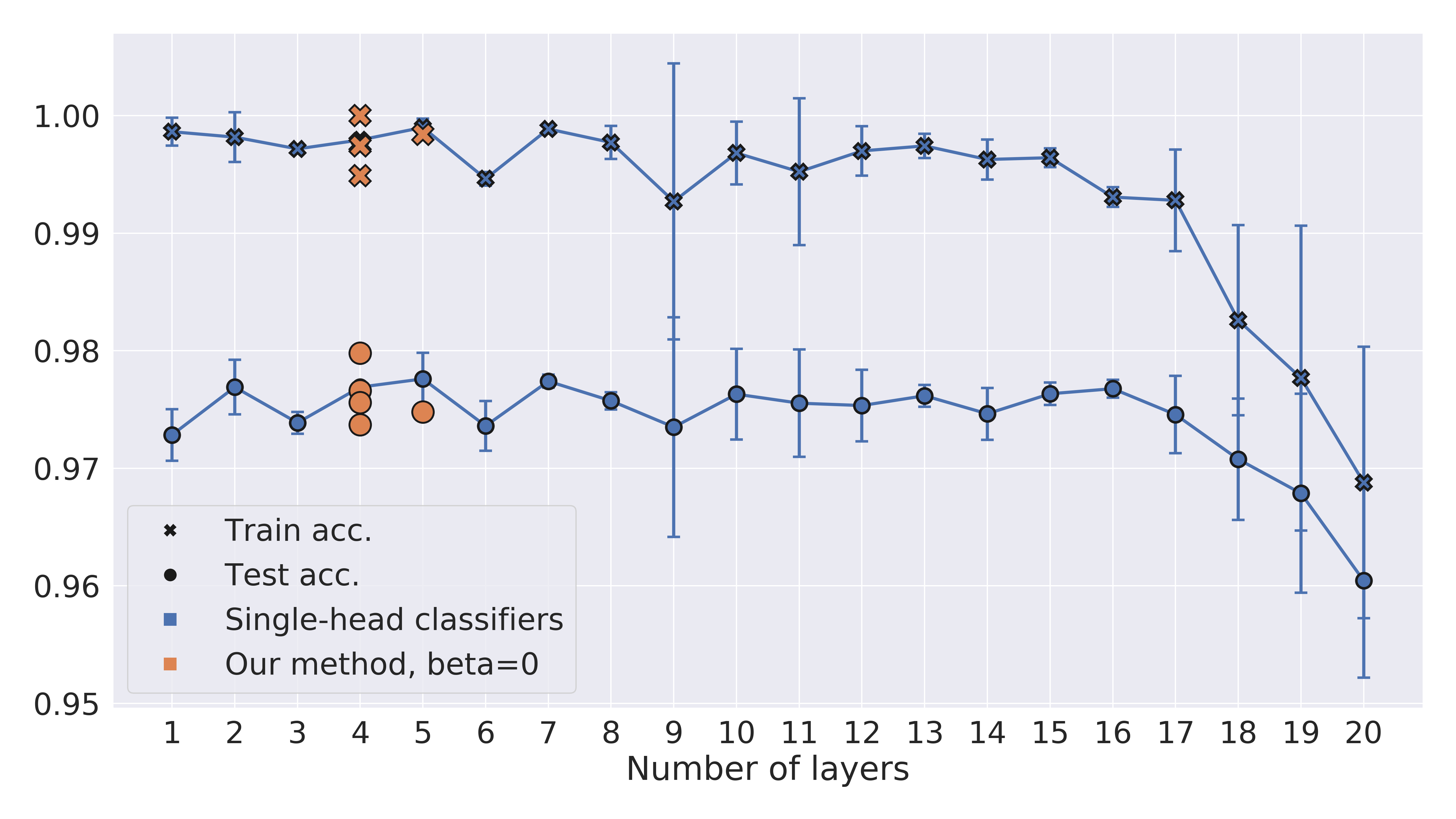}
  (a) MNIST
  \label{fig:baseline_mnist_fc}
\end{minipage}%
\begin{minipage}{.5\textwidth}
  \centering
  \includegraphics[width=\linewidth]{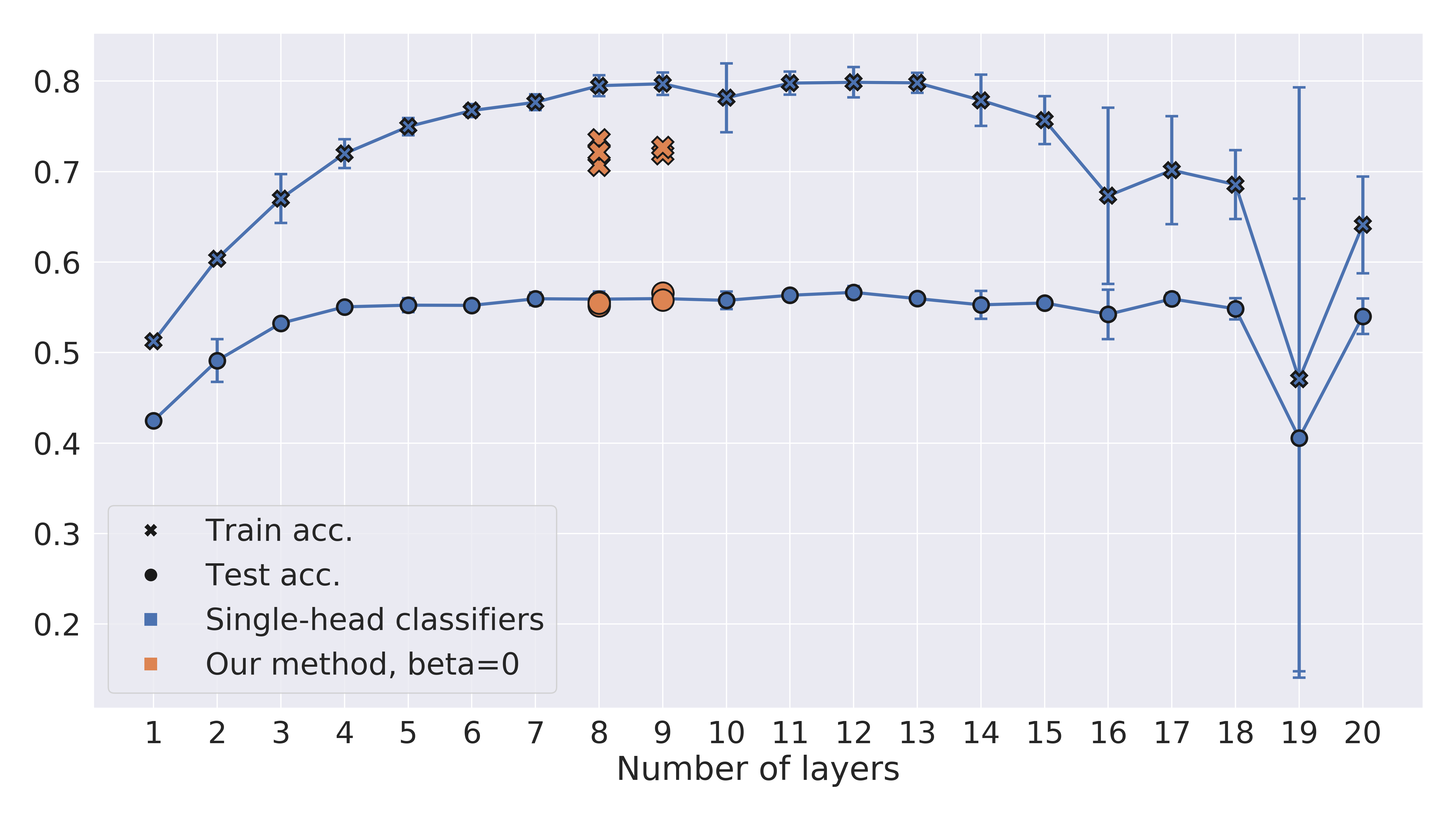}
  (b) CIFAR-10
  \label{fig:baseline_cifar_fc}
\end{minipage}%
\caption{Accuracy of networks resulting from applying our method to FC architectures. For both datasets, our method finds shallower networks with almost equal test accuracy.}
\label{fig:baselines_mnist_cifar_fc}
\end{figure}

Results presented in Fig. \ref{fig:baselines_mnist_cifar_fc} show that the method achieves better test accuracy scores than the base 20-layer network and similar scores to the best network tested, while significantly reducing the computational complexity. We determine the nature of low accuracy in the deeper single-head networks as an effect of gradient instability and explore this problem further in Section~\ref{subsection:vanishinggradient}.

\subsection{Graph-based networks}

\begin{figure}[h!]
\centering 
\begin{minipage}{.37\textwidth}
  \centering
  \includegraphics[width=\linewidth]{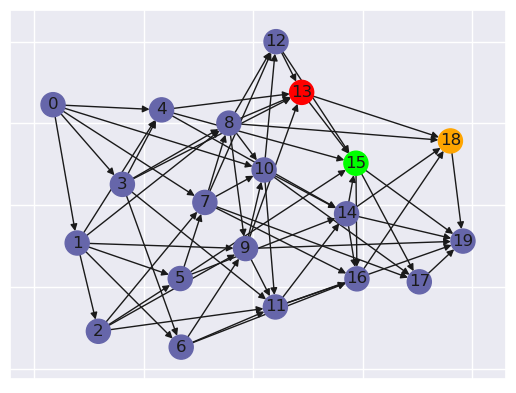} 
  (a)
\end{minipage}\hfill%
\begin{minipage}{.56\textwidth}
    \centering 
    \includegraphics[width=\textwidth]{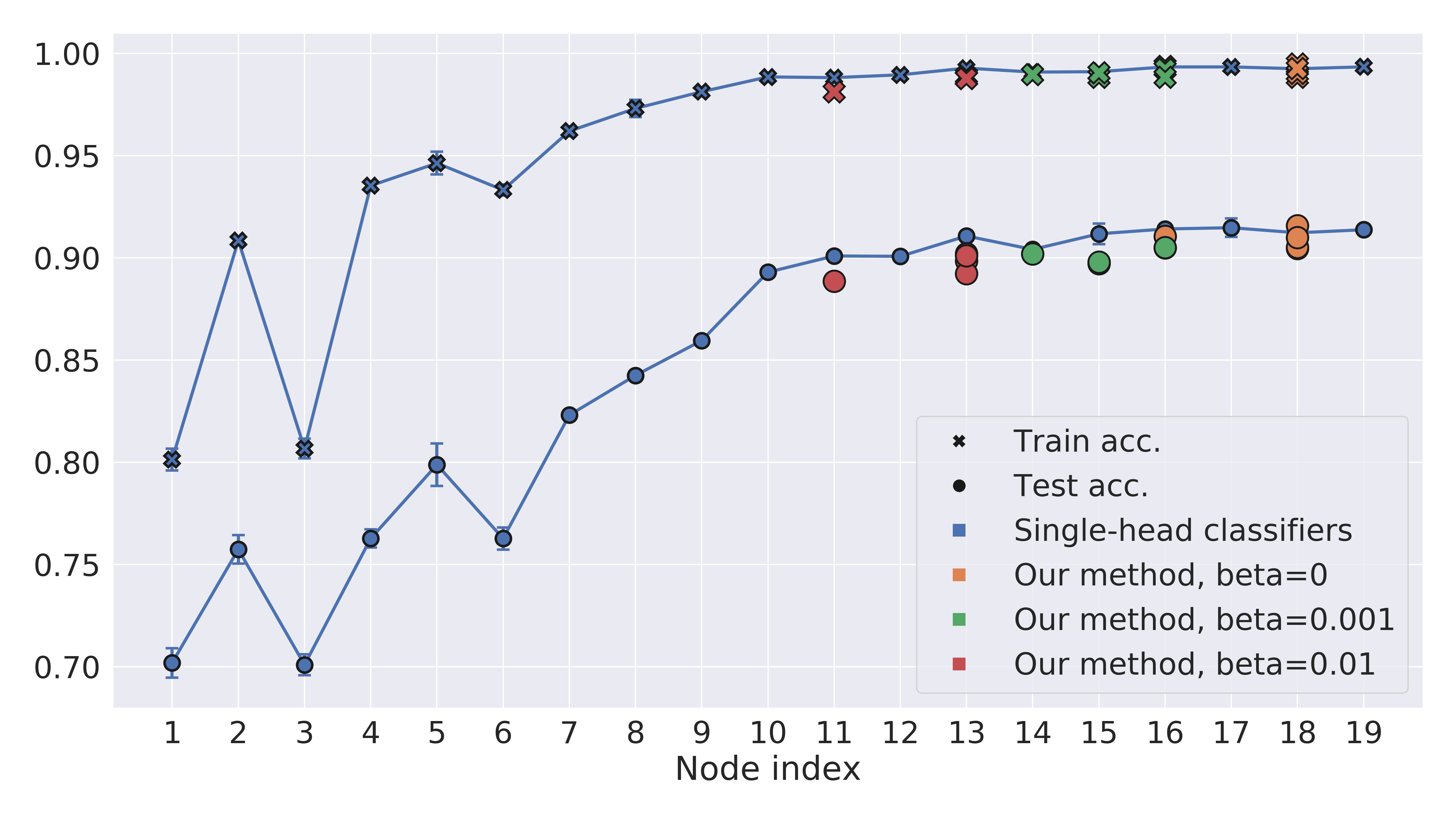}
    (b)
\end{minipage}
  \caption{(a) The graph used for generating the network used in the experiment. Nodes most commonly selected by the model for each $\beta$ are marked with colors corresponding to the legend on the plot on the right. (b)~Accuracy of the network with \our~on CIFAR-10.}
\label{fig:graphs_exp_1}
\end{figure}
\label{sec:graphs}
And lastly, we test \our{} on a more general class of architectures with more complex connection patterns. We generate a random graph and then transform it into a neural network architecture, by treating each node as a ResNet-like block and using outputs of all its predecessors as inputs. This subsection is based on Janik \& Nowak \cite{janik2020neural}, and we refer the reader to their work for details about the network generation process.\footnote{The implementation we are using for this experiment can be found in this GitHub repository: \url{https://github.com/rmldj/random-graph-nn-paper}}

For the first experiment, we use a randomly generated graph shown in Fig. \ref{fig:graphs_exp_1}.
 We add a classifier head to each ResNet-like block (represented by a node in the graph) and run our method with the expectation that we will be able to find the optimal subgraph of the original graph. In order to adapt our method to graphs, we redefine the time-regularization component of the loss function: 
$$
L_{reg} = \sum_k w_k e(k),
$$
where $e(k)$ denotes the number of edges in a subgraph consisting of predecessors of node $k$ and the node itself. The form of this regularization component is again chosen to approximate the time it takes the network to process an example.

Similarly as before, we repeat each run of our method 5 times for each $\beta$ and compare the results to the baseline situation, where each possible subgraph has one classifier head on the final node. The results presented in Fig. \ref{fig:graphs_exp_1} are consistent with the ones from previous experiments, further confirming the stability of \our{}, even for non-standard architectures with complex connection patterns.

\section{Analysis}

\subsection{Optimization properties of the network}
\label{Section:optimization}
The dynamics of training multi-head networks are non-trivial and still not well understood. In order to explore the optimization properties of \our{}, we take inspiration from the multi-task learning literature. To estimate whether the interference between different tasks is beneficial or detrimental, several works compare the directions of gradients of each task \cite{du2018adapting,yu2020gradient}. If the gradients of two tasks point in the same direction, then positive reinforcement occurs. If they point in opposite directions, then tasks are harmful to each other, and for orthogonal directions they are independent.

For our analysis, we use the cosine similarity metric between two vectors:
\begin{equation*}
    \rho(v, u) = \frac{\il{v, u}}{\sqrt{\il{v,v}\il{u,u}}},
\end{equation*}
where $\rho \in [-1, 1]$, with $-1$ appearing for opposite vectors, $0$ for orthogonal vectors, and $1$ for vectors pointing in the same direction.

We would like to measure the impact of each classifier in the network on the parameters. In order to do that, we introduce the notion of a partial gradient with respect to parameters $\theta$, defined as
\begin{equation*}
    \hat{g}^l_\theta = \frac{\partial L(\hat{o}_1, \ldots, \hat{o}_n)}{\partial \hat{o}_l} \frac{\partial \hat{o}_l}{\partial \theta},
\end{equation*}
where $n$ is the number of classifiers.

The partial gradient can be seen as the gradient of the loss function obtained by pushing the gradient only through the $\hat{o}_l$ and stopping it for all $\hat{o}_k$ where $k \neq l$. In practice, $\hat{g}^k_\theta$ represents the change of the parameters $\theta$ expected by the $k$-th classifier.

We can see that the sum of all the partial gradients gives us the "full" gradient:
\begin{equation*}
    \sum_k \frac{\partial L(\hat{o}_1, \ldots, \hat{o}_n)}{\partial \hat{o}_k} \frac{\partial \hat{o}_k}{\partial \theta} = \nabla_\theta L(\hat{o}_1, \ldots, \hat{o}_n) \eqdef \hat{g}_\theta.
\end{equation*}

\begin{figure}[h!]
  \centering
    \begin{minipage}[b]{.67\textwidth}
      \centering
      \includegraphics[width=\linewidth]{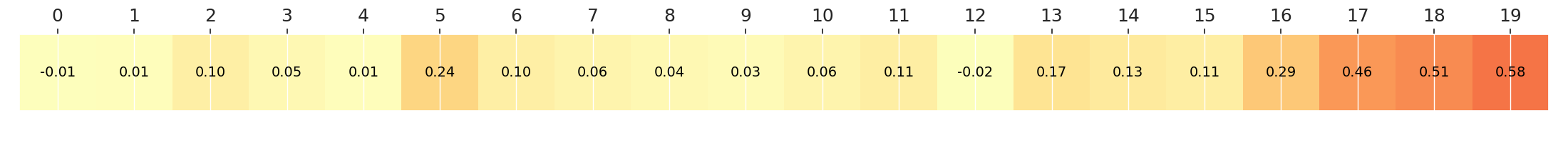}
      \includegraphics[width=\linewidth]{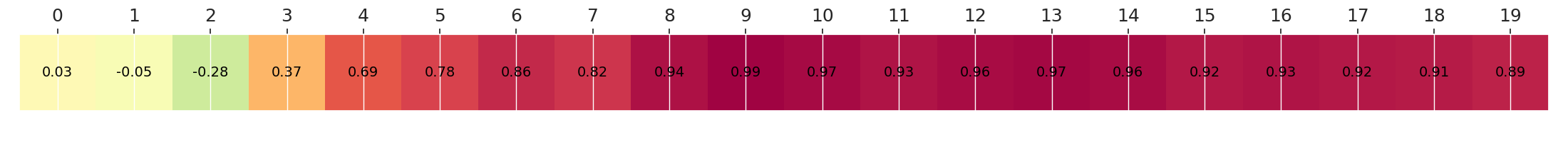}
      \includegraphics[width=\linewidth]{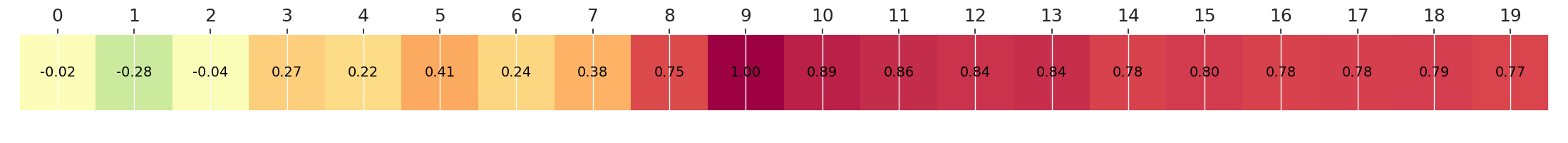}
      \includegraphics[width=\linewidth]{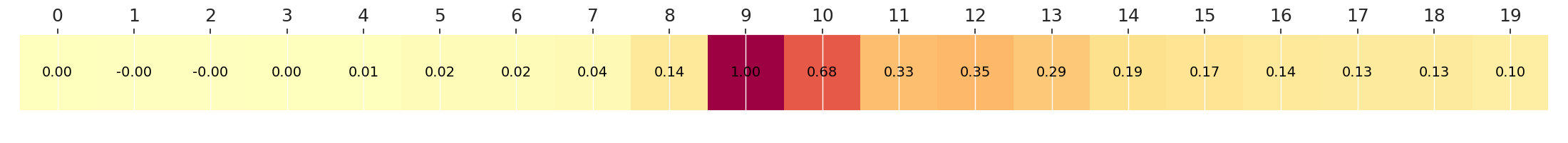}
      (a)
  \end{minipage}%
  \begin{minipage}[b]{.32\textwidth}
    \centering
    \includegraphics[width=\linewidth]{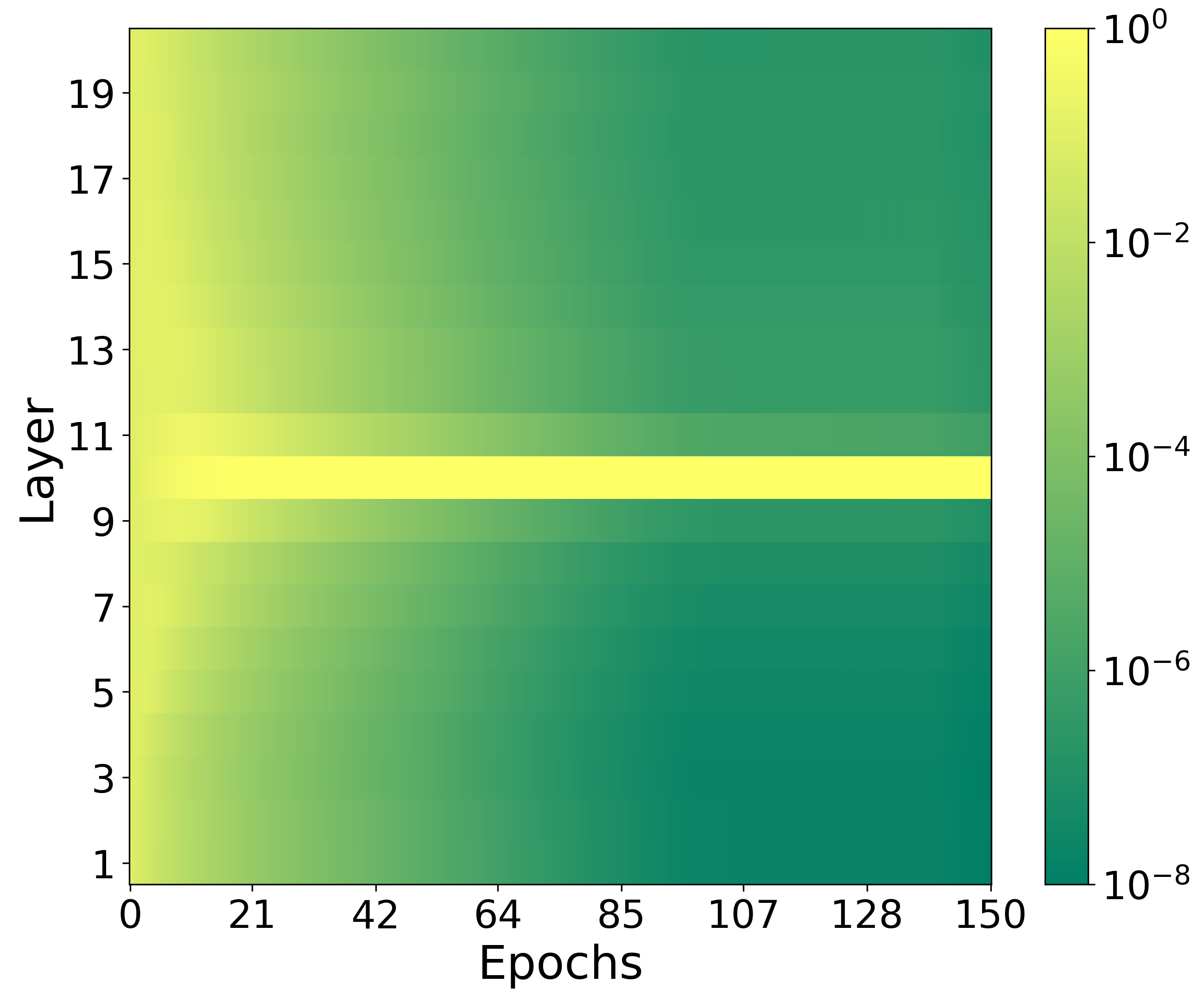}
    (b)
    \label{fig:cs_weights}
  \end{minipage}%
  \label{fig:true_cs}
  \caption{(a) Cosine similarities of the full gradient and partial gradients, calculated with respect to the first layer of the network. The subsequent vectors, starting from the top, were recorded at the beginning of the training, the 10th, the 45th and the 140th epoch, respectively. We see that after the training stabilizes, the partial gradient of the chosen layer is most directly correlated with the true gradient. (b) The head importance weights of the network throughout the training process.}
\end{figure}

\begin{figure}[h!]
\centering 
\begin{minipage}{.5\textwidth}
  \centering
  \includegraphics[width=\linewidth]{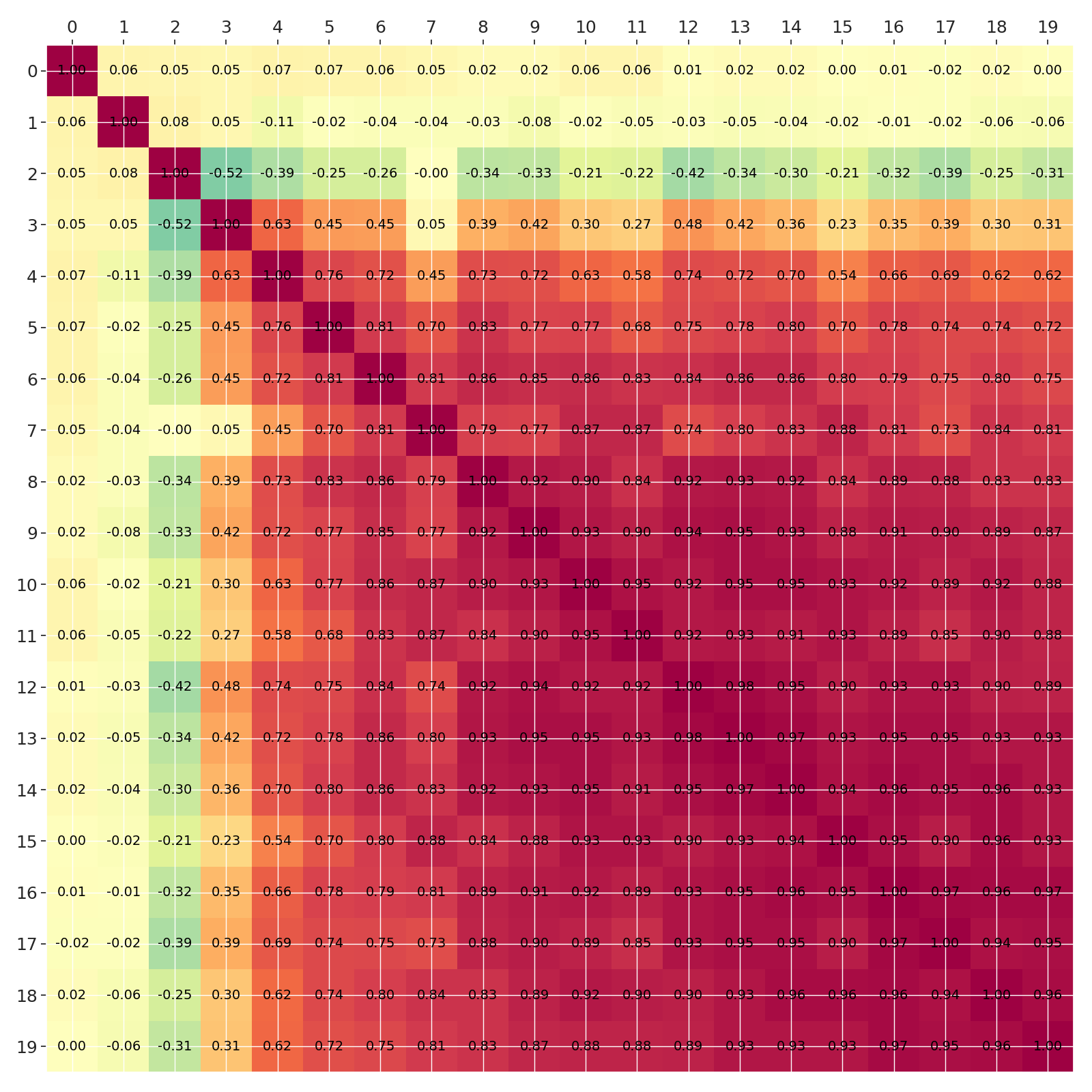}
  (a)
  \label{fig:cs_matrix_early}
\end{minipage}%
\begin{minipage}{.5\textwidth}
  \centering
  \includegraphics[width=\linewidth]{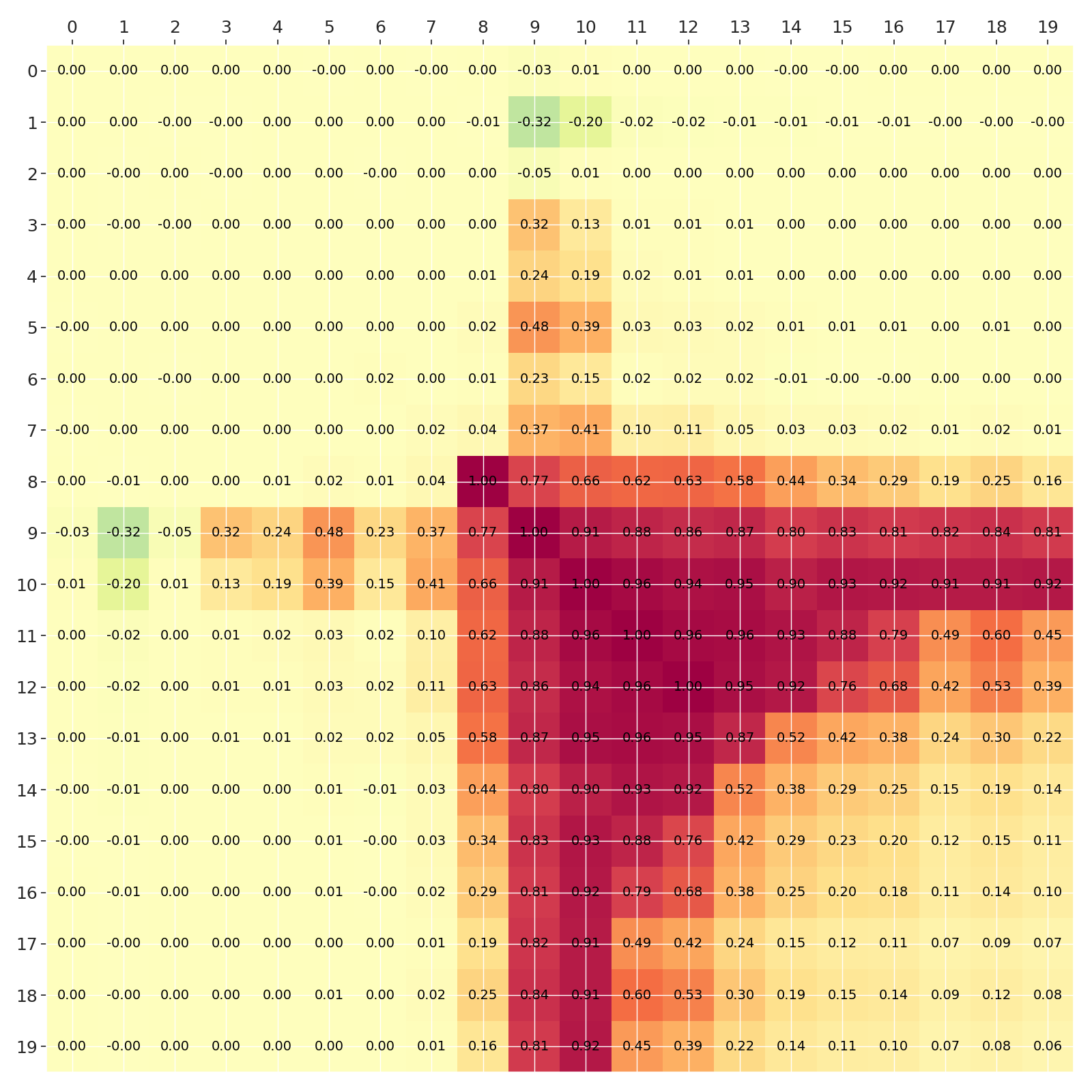}
  (b)
  \label{fig:cs_matrix_middle}
\end{minipage}%
\caption{Cosine similarities of the partial gradients with respect to the first layer after (a) 10 epochs and (b) 45 epochs. The cell with coordinates $(x, y)$ shows the value of cosine similarity between respective partial gradients $\rho(\hat{g}_\theta^{x}, \hat{g}_\theta^{y})$.
We can see that the gradients of the later classifier heads are significantly more correlated than gradients of the earlier ones.
}
\label{fig:cs_matrices}
\end{figure}
 
To better understand how different classifier heads affect the learning process, we compute cosine similarities between partial gradients and the full gradient throughout the training. Results presented in Fig. \ref{fig:true_cs} show that as the training progresses, the cosine similarity values for the 9th layer, which is the final choice of the model, are consistently increasing. This is because when the head's importance weight increases, so does the magnitude of its gradients and its impact on the full gradient.

Another interesting phenomenon appears in the earliest stages of training. At the very first step, most of the partial gradients are almost orthogonal to the full gradient, which suggests high levels of noise at the beginning of training. However, during the first few epochs, the cosine similarities of the later layers increase significantly, signaling that they agree with each other in large part. This is not the case for the earliest layers, as their partial gradients stay orthogonal to or even point in the opposite direction of the full gradient.

This observation shows why the importance weights of early classifier heads drop so quickly. Since the partial gradient of the early layers is usually orthogonal to the full gradient used for updating the parameters, those classifiers do not have an opportunity to learn any features directly relevant for classification, and thus their influence is quickly reduced.

To further investigate this issue, we check the cosine similarities between each pair of partial gradients $\hat{g}^k$ and $\hat{g}^l$ during training and present the results in Fig. \ref{fig:cs_matrices}. We observe that the cosine similarity in the later parts of the network is much higher than in the earlier parts of the network. Our explanation for this phenomenon is that the earlier layers strive to detect complex features that would enable proper classification even in a shallow network, while all of the later layers expect the base of the network to learn the same set of simple patterns. This finding seems to be consistent with the well-known phenomenon of early layers functioning as basic filters for detecting small local patterns such as edges, with only the deeper layers encoding more complex structures.

\subsection{Poor starting conditions}
\label{subsection:vanishinggradient}

\begin{figure}[h!]
\centering 
\includegraphics[width=0.7\textwidth]{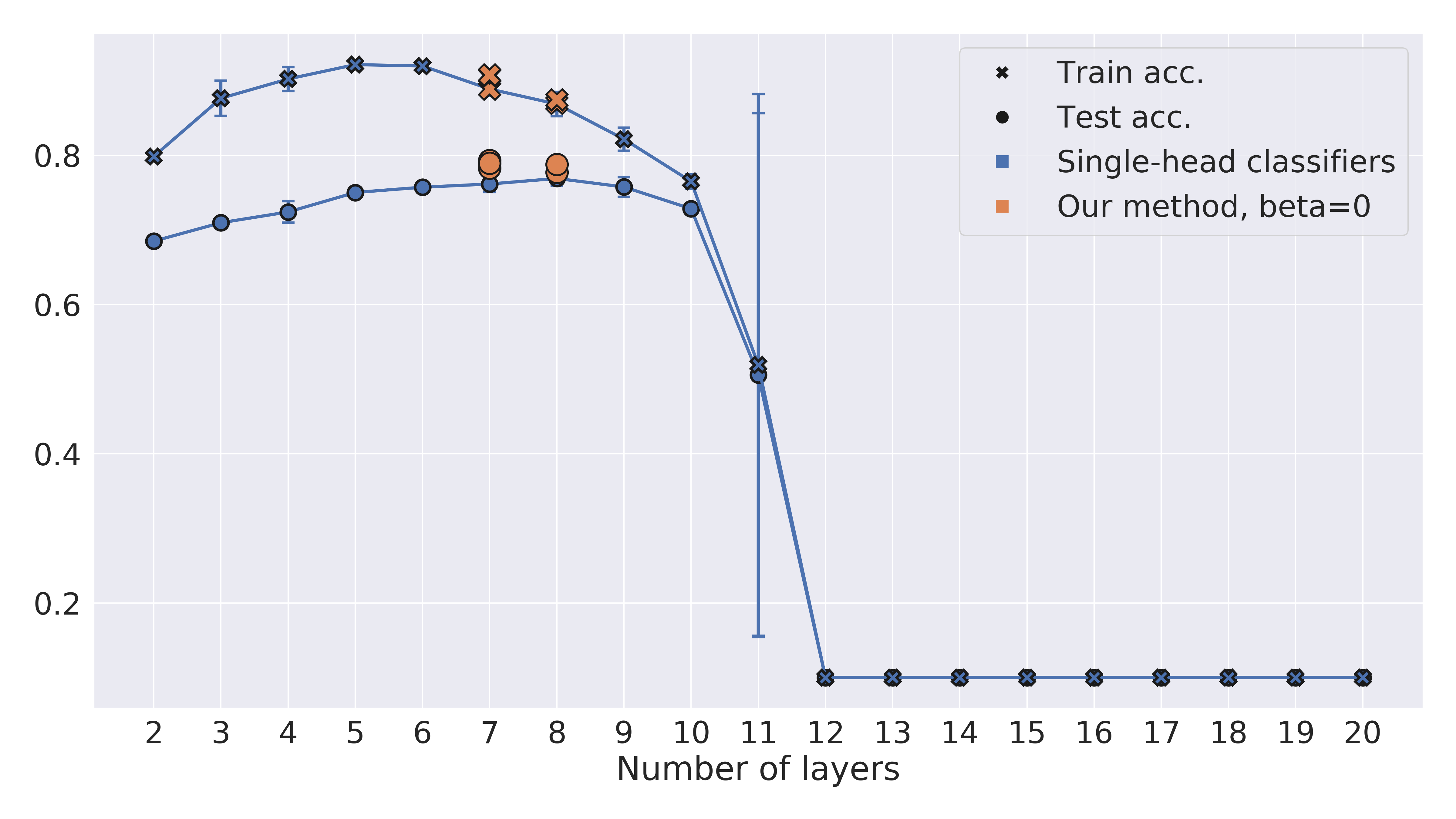}
\caption{Accuracy of \our{} for models that exhibit vanishing or exploding gradient problems. We see that even if the starting network with $20$ layers collapses, our method is able to find a well-performing subnetwork.}
\label{fig:vanishing_gradient}
\end{figure}

An interesting question is how \our{} performs in situations where the starting network fails at the given task, but contains a subnetwork that performs well. To investigate this point, we construct a setting where gradient vanishing or exploding phenomena occur. When the network is too deep and proper optimization techniques are not used, the noise in the gradients accumulating during the backward step will make it impossible to learn. However, if we were to use only a subset of layers of this network, such a model would be then able to achieve satisfactory performance.

To test this, we use our CNN architecture introduced in the experiments section, but without the batch normalization layers, which are known for their effect on reducing the gradient instability \cite{santurkar2018does}. The results presented in Fig. \ref{fig:vanishing_gradient} show that networks with more than 10 layers are highly unstable, with their final performance being no better than random.

At the same time, our network is able to pick a subnetwork consisting of 7 layers, remaining in the region where the training is still stable and thus achieving good performance. This result suggests that \our{} is able to counteract the effect of choosing a poor starting network for the given task.

\subsection{Analysis of head importance weights behaviour}
\begin{figure}[h!]
\centering 
\begin{minipage}[t]{.45\textwidth}
  \centering
  \includegraphics[height=3.9cm,keepaspectratio]{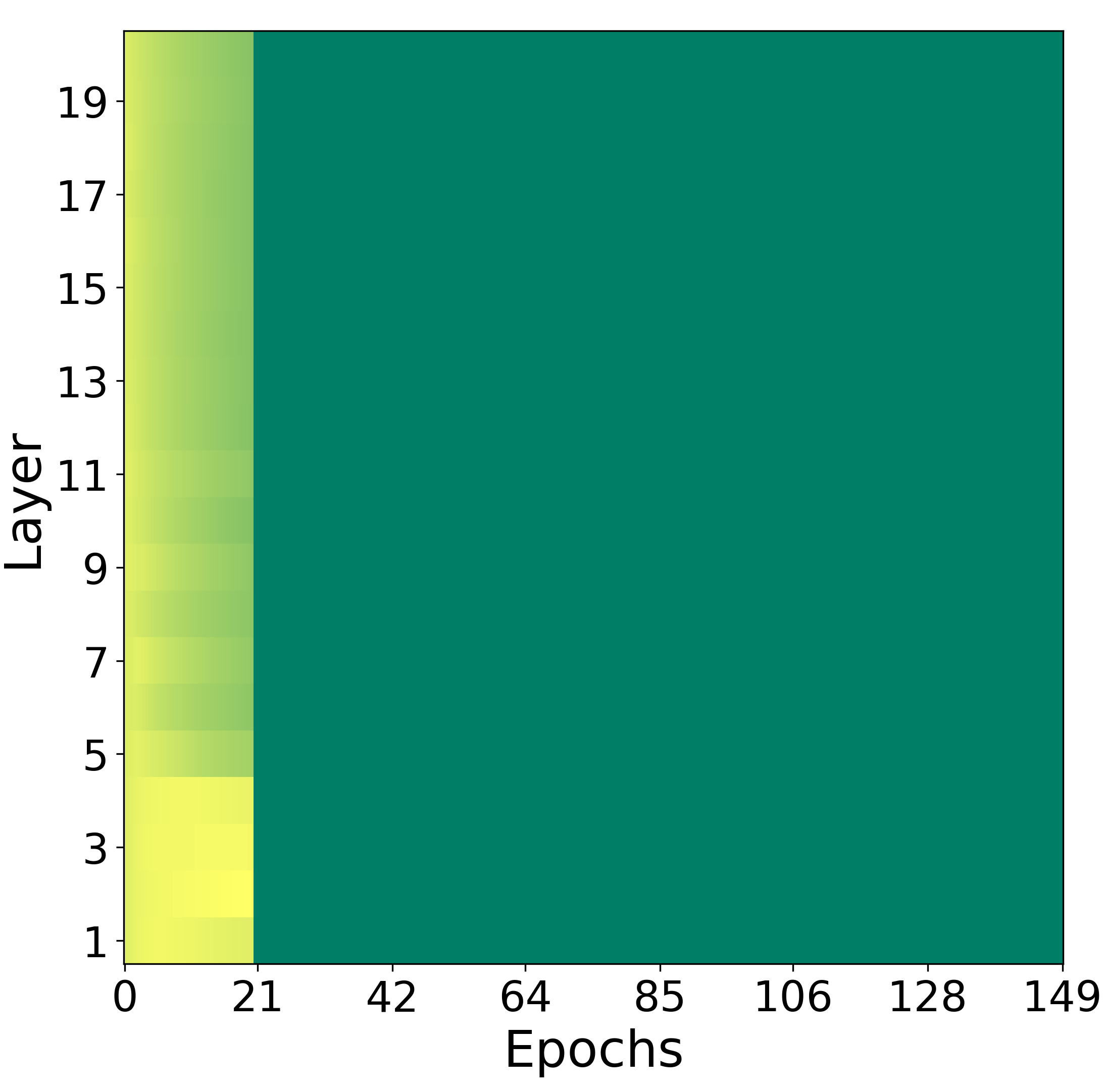} \\
  (a) Combining probabilities (softmax)
\end{minipage}%
\begin{minipage}[t]{.45\textwidth}
  \centering
  \includegraphics[height=3.9cm,keepaspectratio]{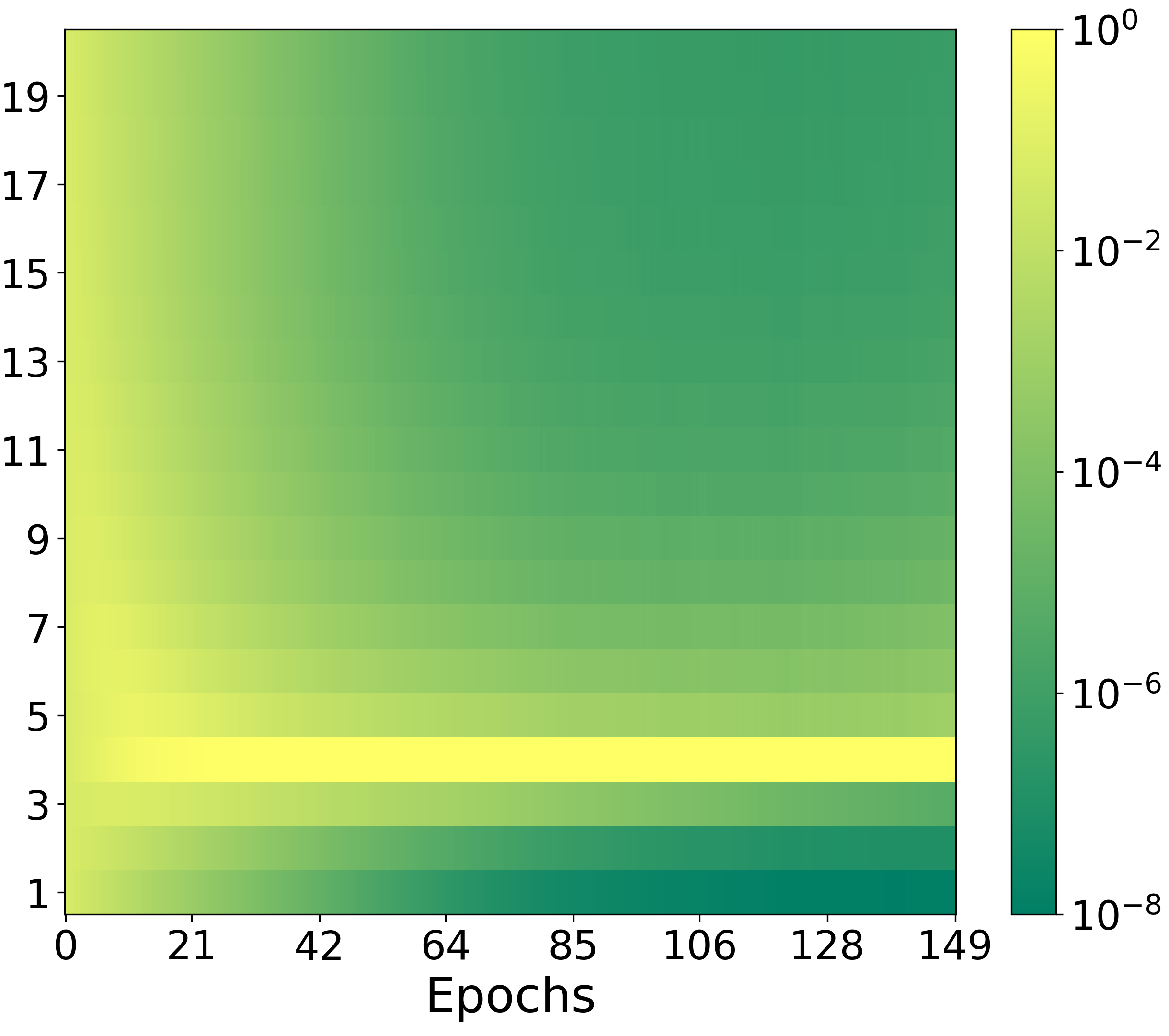} \\ 
  (b) Combining log probabilities (log-softmax)
\end{minipage}%
\caption{Two different ways of producing the final output for the network. The dark green values after the 21st epoch on the first plot represent NaN values appearing after overflow occurrence. Directly combining the probabilities does not guarantee that the model will converge on a single layer and can lead to numerical overflows.}
\label{fig:softmax_vs_log}
\end{figure} 

\label{subsection:weightnormalization}

\begin{figure}[h!]
\centering 
\begin{minipage}{.32\textwidth}
  \centering
  \includegraphics[height=3.9cm,keepaspectratio]{img/beta_00_cifar_conv}
  (a) uniform (baseline)
  \label{fig:init_uniform}
\end{minipage}%
\begin{minipage}{.33\textwidth}
  \centering
  \includegraphics[height=3.9cm,keepaspectratio]{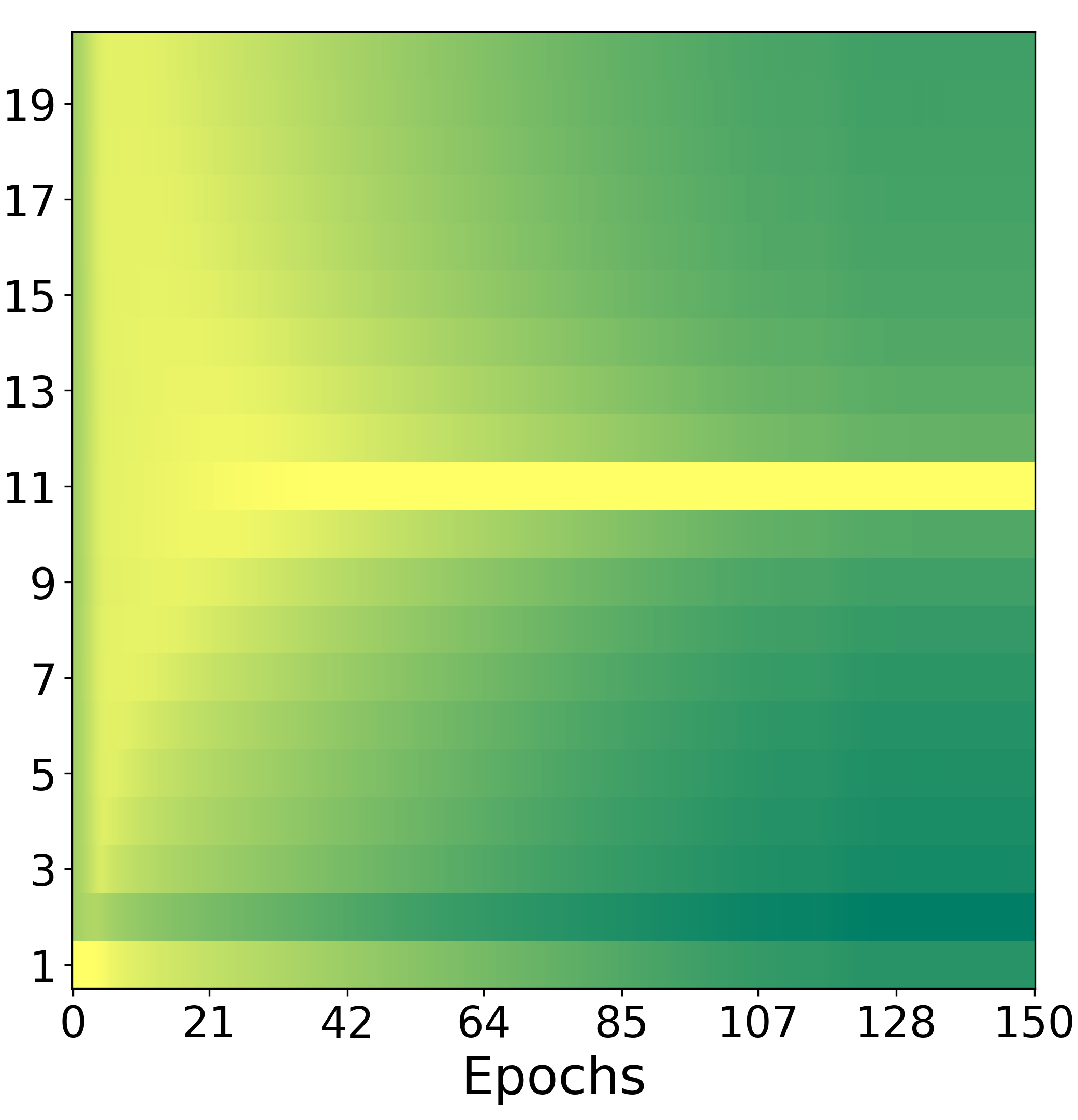}
  (b) first
  \label{fig:init_first}
\end{minipage}%
\begin{minipage}{.33\textwidth}
  \centering
  \includegraphics[height=3.9cm,keepaspectratio]{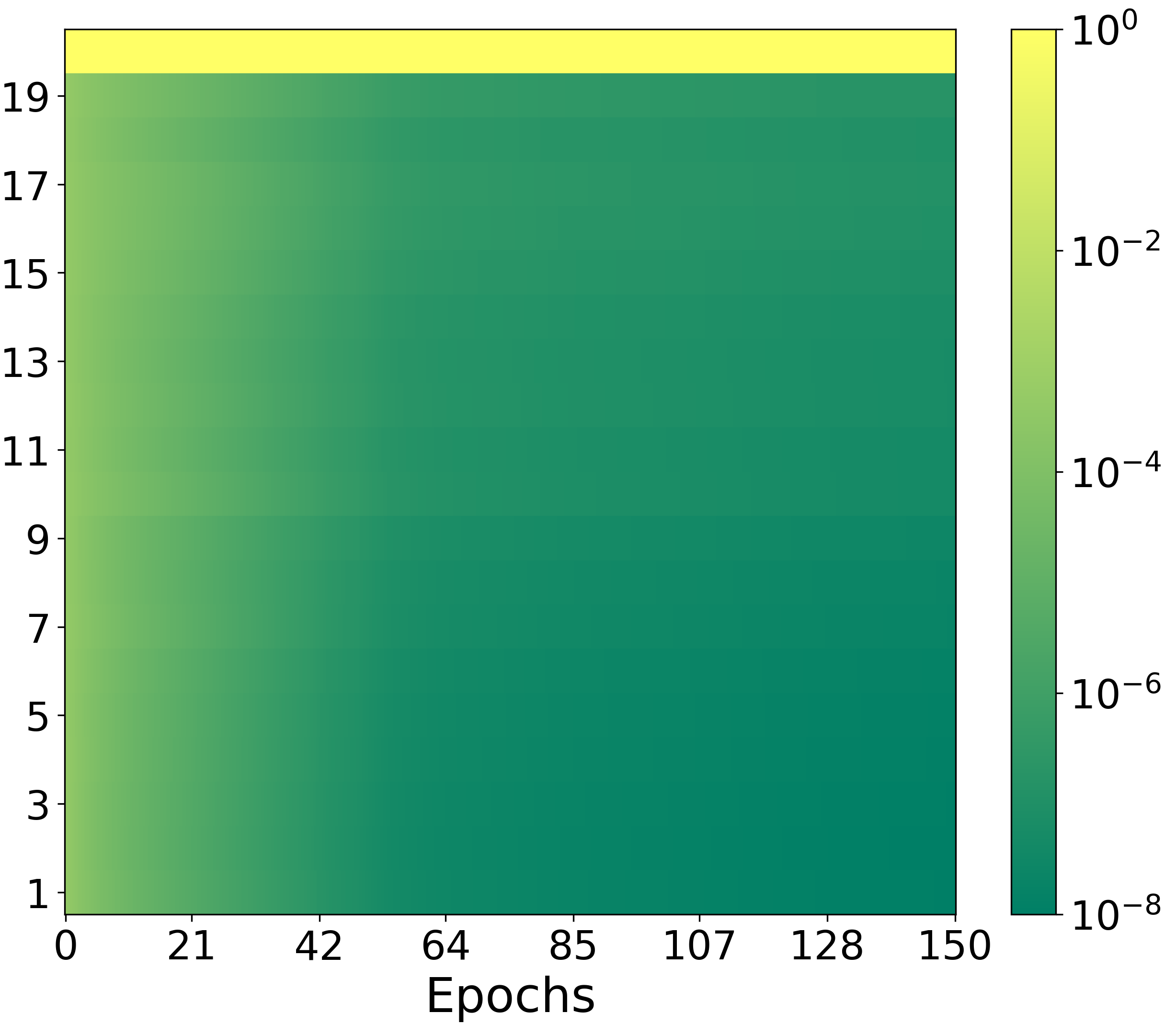}
  (c) last
  \label{fig:init_last}
\end{minipage}%
\caption{Head importance weights progress with alternative init schemes with $\beta=0$. Although the network behaves the same for uniform and first-head initializations, it stays at the last layer when initialized so. The achieved accuracy scores are 86.47\%, 85.94\%, and 87.98\% for uniform, first, and last schemes, respectively.}
\label{fig:init_figures}
\end{figure}

Head importance weights $w_k$ used for weighting the output of each classifier decide the shape of the final obtained network and as such are crucial for our method. To better understand the function they perform in \our{}, we check their properties in different settings.

We examine the properties of the log softmax trick in \our{} by investigating the baseline situation where for the final prediction we combine the probabilities themselves instead of the logarithm of the probabilities. We compare these two approaches on a fully-connected network trained on MNIST and present the results in Fig. \ref{fig:softmax_vs_log}. As we can observe, combining the probabilities themselves leads to divergence of the training due to overflow happening in the softmax function, which is required for obtaining the probabilities. Similar issues do not appear in our approach since we can directly calculate the logarithm of the softmax operation, which is much more stable numerically.

Another important difference to note is that the baseline approach of combining probabilities does not lead to the choice of a single classifier, i.e. there are multiple head importance weights $w_k$ with values significantly larger than zero. For such a model, cutting out a single-head classifier would lead to significant performance drop, because of the lack of influence of the removed heads.

To further understand the optimization properties of \our{}, we examine the behavior of our method for different initialization schemes for importance head weights $w_k$. In the standard approach we initialize the weights uniformly. Here, we test two additional variants that put almost the entire weight onto the first or the last classification head. Results presented in Fig. \ref{fig:init_figures} show that the first-head initialization does not differ significantly from our standard approach. However, starting with most of the probability mass on the last head, the network does not move the mass back to the earlier layers.

\section{Conclusion}
We presented \our{}, a simple end-to-end method for compressing and accelerating neural networks, which improves the inference speed both on GPU and CPU. As part of our method, we introduced a novel way of combining outputs of multiple classifiers. Combining log probabilities instead of probabilities themselves allows us to converge to a single layer and avoid numerical instabilities connected with computing the softmax function. We show extensive analysis of the optimization properties of the network, including the investigation of gradient directions inspired by multi-task learning methods.


%
%
%
\bibliographystyle{splncs04}
\bibliography{ref}
%
\end{document}